\documentclass{article}

\usepackage{PRIMEarxiv}

\usepackage[utf8]{inputenc} 
\usepackage[T1]{fontenc}    
\usepackage{hyperref}       
\usepackage{url}            
\usepackage{booktabs}       
\usepackage{amsfonts}       
\usepackage{nicefrac}       
\usepackage{microtype}      
\usepackage{lipsum}
\usepackage{fancyhdr}       
\usepackage{graphicx}       
\graphicspath{{media/}}     
\usepackage{xcolor}
\usepackage{amsmath}
\usepackage{times}
\usepackage{latexsym}
\usepackage{amsfonts}
\usepackage{graphicx}
\usepackage{caption}
\usepackage{subcaption}
\usepackage{arydshln}
\usepackage{natbib}
\usepackage[utf8]{inputenc}
\usepackage[english]{babel}
\title{Aspect and Opinion Term Extraction Using Graph Attention Network
}

\author{
  Abir Chakraborty \\
  Microsoft \\
  \texttt{Abir.Chakraborty@microsoft} \\
}

\begin{document}
\maketitle

\begin{abstract}
In this work we investigate the capability of Graph Attention Network for extracting aspect and opinion terms. Aspect and opinion term extraction is posed as a token-level classification task akin to named entity recognition. We use the dependency tree of the input query as additional feature in a Graph Attention Network along with the token and part-of-speech features. We show that the dependency structure is a powerful feature that in the presence of a CRF layer substantially improves the performance and generates the best result on the commonly used datasets from SemEval 2014, 2015 and 2016. We experiment with additional layers like BiLSTM and Transformer in addition to the CRF layer. We also show that our approach works well in the presence of multiple aspects or sentiments in the same query and it is not necessary to modify the dependency tree based on a single aspect as was the original application for sentiment classification.
\end{abstract}


\section{Introduction}
Extracting information from customer feedback is a key capability required for identifying current drawbacks and scope for further improvement. Online shoppers routinely provide feedback on their experience with the purchased product that are not just important for the other potential customers but also a critical feedback to the product manufacturers for the next cycle of iteration. Similar feedbacks are available in various other domains ranging from Manufacturing to Healthcare where granular opinions (sentiments) about various dimensions (aspects) of the used product (or service) are available in textual form but need to be understood. Due to the presence of multiple aspects (and the corresponding sentiments) extraction of these aspect-sentiment pairs is a challenging task and since its introduction in 2014 (SemEval-2014 Task-4, \citeauthor{pontiki-etal-2014-semeval}) Aspect Based Sentiment Analysis (ABSA) attracted various different approaches and is still under active consideration. 

ABSA demands semantic understanding of the sentence where it is necessary to identify the aspect terms (defining "what") and the opinion terms (defining "why") and the connection between each related pairs (resulting in a positive, neutral or negative sentiment, i.e., "how"). A relatively simple example would be “The price is reasonable although the service is poor” where "price" and "service" are the aspects with the corresponding opinion terms "reasonable" and "poor", respectively. Here, the relative locations of the aspect and opinion word can help each other to identify these terms. However, other examples like “prices are in line” (neutral sentiment) and “For the price, you can not eat this well in Manhattan” (positive sentiment) it is not obvious which opinion terms are driving the sentiments and their linkage with the aspect terms. As a result, while it is possible to capture the syntactical structure and dependency between words it is necessary to capture a deeper meaning of each token (word) that should not be relied upon the limited number of examples (that are hallmark of all ABSA datasets) but rich representations provided by large language models like BERT and RoBERTA. 

\begin{table*}
\centering
\begin{tabular}{llcccc}
\hline
\textbf{Dataset} & & Train & Dev & Test & Total \\
\hline
LAPTOP & \#sent & 2741 & 304 & 800 & 4245 \\
      & \#aspect & 2041 & 256 & 634 & 2931 \\
\hline
REST & \#sent & 3490 & 387 & 2158 & 6035 \\
   & \#aspect & 3893 & 413 & 2287 & 6593 \\
\hline
\end{tabular}
\caption{Statistics of the dataset in \cite{li-etal-2019-exploiting}}
\label{astev0_data}
\end{table*}

In addition to large model based initial representations, additional information from parts-of-speech and dependency structure play a critical role as can be seen in some of the previous ABSA work where the objective is to predict only the sentiment associated with a particular aspect. While it is generally understood that a confluence of deep encoders and graph-based representation of the input sentence will drive better performance it is not clear what is the optimal graph representation of a sentence, especially for ABSA kind of task. In case of polarity detection, it can be argued that only the part of the dependency tree associated with a particular aspect is significant and the rest of the tree can be ignored. However, when we convert the task into aspect and opinion term extraction it is not clear whether any part of the sentence has higher significance compared to the rest. 

In this work we address the problem of aspect and opinion term extraction from input sentences. As an example, "the \textcolor{blue}{weather} was \textcolor{red}{gloomy}, but the \textcolor{blue}{food} was \textcolor{red}{tasty}" has two sets of aspect-opinion-sentiment tuples, i.e., (\textcolor{blue}{weather}, \textcolor{red}{gloomy}, negative) and (\textcolor{blue}{food}, \textcolor{red}{tasty}, positive). We use ABSA datasets with token level labels - one set for the aspects and one set for the opinions. Thus, the aspect and opinion term extraction becomes a NER task with different types of entity classes. The tags for the aspects include the sentiment, (for opinion it is only the BIEOS tag), however, what is missing is the connection between the aspects and opinion terms. Thus, our aspect models predict (\textcolor{blue}{weather}, negative) and (\textcolor{blue}{food}, positive), opinion models predict (\textcolor{red}{gloomy}) and (\textcolor{red}{tasty}), without making the subsequent contractions. We encode the graph associated with the dependency parsing of the input sentence and create token (node) level representations that take care of both the neighborhood (connected edges) and dependency type (edge type) of each token.

The organization of the paper is as follows. In the next section we provide a detailed literature survey on the techniques employed for aspect and opinion extraction task of ABSA. Next, we present the details of the proposed model. Subsequently, the model predictions and comparisons with other baseline methods are discussed. Finally, conclusions are drawn and scope for future works is outlined.

\section{Related Work}
Most of the ABSA work concern with the sentiment polarity detection associated with a particular aspect and the approaches varied from initial SVM classifier with handcrafted features to deep learning classifiers based on RNN \citep{wang-etal-2016-attention, ijcai2018-617, he-etal-2018-effective, ijcai2017-568}, Transformer \citep{hoang-etal-2019-aspect, app9163389, xu2019bert} and memory network \citep{tang-etal-2016-aspect, chen-etal-2017-recurrent, 10.1145/3132847.3132936, ijcai2019-707}. There are few graph based approaches as well, e.g., graph convolution network (GCN) based model of \citet{zhang-etal-2019-aspect} and \citet{sun-etal-2019-aspect}, graph attention network (GAT) based model of \citet{huang-carley-2019-syntax} and \citet{wang-etal-2020-relational} where the latter modified the original dependency tree to create an aspect-oriented dependency tree that was used further in a relational GAT (R-GAT) where different relations contributed differently in the computation of nodal representations. However, the approach assumes the presence of only one aspect at a time in the given input.

On the other hand, an end-to-end ABSA tries to extract all the aspect terms in a given query simultaneously along with the corresponding sentiments. The first approach towards E2E-ABSA was provided by \citet{li-etal-2019-exploiting} where a unified tagging scheme (combining the position and sentiment) was used, i.e., the token labels were one of {\tt B-\{POS,NEG,NEU\}}, {\tt I-\{POS,NEG,NEU\}},
{\tt E-\{POS,NEG,NEU\}}, {\tt S-\{POS,NEG,NEU\}} or {\tt O}, denoting the beginning, inside or end of an aspect, single-word aspect, with positive, negative or neutral sentiment respectively, and finally the outside of that aspect. In this work, BERT \citep{devlin-etal-2019-bert} was used to embed the tokens and other layers like, GRU, Transformer \citep{vaswani2017attention} or CRF was used on the BERT output.  

\citet{li-etal-2019-exploiting} ignored the opinion term extraction and that was addressed by \citet{Peng_Xu_Bing_Huang_Lu_Si_2020} who proposed a two-stage pipeline framework. In the first stage, aspect-sentiment pairs were extracted using the above mentioned unified tagging scheme. In addition, opinion spans were extracted using {\tt BIEOS} tagging scheme. The aspect-sentiment and opinion pairs thus extracted were matched against each other in the second stage where an MLP-based classifier was used to find the compatibility of these pairs. \citet{zhang-etal-2020-multi-task} proposed a multi-task framework to jointly detect aspects, opinions, and sentiment dependencies. However, instead of using the unified tagging scheme they used two sets of {\tt BIOS} tags to identify the aspects and sentiments before connecting them with the corresponding sentiment. While all these approaches extract aspects and opinion pairs in isolation a recent approach based on pointer network is proposed by  \citet{mukherjee-etal-2021-paste} where an encoder-decoder architecture is used to generate all the aspect-opinion-sentiment tuples. 

\begin{table*}
\centering
\begin{tabular}{rccccccc}
\hline
\textbf{Dataset} & \multicolumn{2}{c}{Train} & \multicolumn{2}{c}{Dev} & \multicolumn{2}{c}{Test} & Total \\
& \#s & \#p & \#s & \#p & \#s & \#p & \#s \\
\hline
LAPTOP'14 & 920 & 1265 & 228 & 337 & 339 & 490 & 1487 \\
REST'14 & 1300 & 2145 & 323 & 524 & 496 & 862 & 2119 \\
REST'15 & 593 & 923 & 148 & 238 & 318 & 455 & 1059 \\
REST'16 & 842 & 1289 & 210 & 316 & 320 & 465 & 1372 \\
\hline
\end{tabular}
\caption{\label{astev1_data}Statistics of the dataset ASTE-V1 as given in  \cite{peng-et-al-2020-knowing} (\#s and \#p denote the number of sentences and aspect-opinion pairs, respectively.}
\label{tab:dataset2}
\end{table*}

\begin{table*}
\centering
\begin{tabular}{lr|ccc|ccc}
\hline
\multicolumn{2}{c}{Model} & \multicolumn{3}{c}{Laptop} & \multicolumn{3}{c}{Restaurant} \\
 & & P & R & F1 & P & R & F1 \\
\hline
\hline
& \cite{li-et-al-2019-unified} & 61.27 & 54.89 & 57.90 & 68.64 & 71.01 & 69.80 \\
Existing Models & \cite{luo-etal-2019-doer} & - & - & 60.35 & - & - & 72.78 \\
& \cite{he-etal-2019-interactive} & - & - & 58.37 & - & - & - \\
\hline
& \cite{lample-etal-2016-neural} & 58.61 & 50.47 & 54.24 & 66.10 & 66.30 & 66.20 \\
LSTM-CRF & \cite{ma-hovy-2016-end} & 58.66 & 51.26 & 54.71 & 61.56 & 67.26 & 64.29 \\
& \cite{liu-et-al-2018-empower} & 53.31 & 59.40 & 56.19 & 68.46 & 64.43 & 66.38 \\
\hline
& BERT+Linear & 62.16 & 58.90 & 60.43 & 71.42 & 75.25 & 73.22 \\
& BERT+GRU & 61.88 & 60.47 & 61.12 & 70.61 & 76.20 & 73.24 \\
BERT Models & BERT+SAN & 62.42 & 58.71 & 60.49 & 72.92 & 76.72 & \underline{74.72} \\
 \cite{li-etal-2019-exploiting} & BERT+TFM & 63.23 & 58.64 & 60.80 & 72.39 & 76.64 & 74.41 \\
& BERT+CRF & 62.22 & 59.49 & 60.78 & 71.88 & 76.48 & 74.06 \\
\hline
Our Model & RGAT-BERT & 62.17 & 60.08 & 61.11 & 69.70 & 73.29 & 71.45 \\
 & RGAT-BERT-CRF & 64.72 & 59.09 & 61.78 & 73.15 & 69.82 & 71.45 \\
 & RGAT-BERT-BiLSTM-CRF & 65.34 & 61.46 & \textbf{63.34} & 75.19 & 71.89 & 73.48 \\
 & RGAT-BERT-TRFMR-CRF & 65.03 & 60.28 & \underline{62.56} & 80.16 & 74.01 & \textbf{76.96} \\
\hline
\end{tabular}
\caption{\label{astev0_results}Comparison of predictions on the aspect extraction dataset of \cite{li-etal-2019-exploiting}. The best F1-scores are shown in bold and the second best ones are underlined.}
\label{tab:perf-astev0}
\end{table*}

\section{Methodology}
We have modified the R-GAT approach of \cite{wang-etal-2020-relational} which was originally targeted for polarity prediction. The approach utilizes the dependency structure of the input sentence which captures the grammatical structure by connecting the words with the corresponding dependency type. The limitations of common approaches that do not pay attention to the parts-of-speech and the dependency structure are discussed by \citet{wang-etal-2020-relational} and the importance of the syntactic relations are emphasized, especially in the context of the aspect word. 

\subsection{Aspect Oriented Dependency Tree}
To bring out the relations of different words with the aspect word(s) an Aspect Oriented Dependency Tree (AODT) was proposed where the root of the original dependency tree was shifted to the target aspect word followed by pruning some of the unnecessary relations. It is to be noted that this approach assumes that there is no more than one aspect word (with the possibility of multiple sub-words) in the sentence. As a result, this approach cannot be applied directly for sentences with multiple aspect words and sentences with no identified aspect word (say, for opinion term extraction).

\subsection{Relational Graph Attention Network}
AODT can be represented by a graph structure where each node is a word and the edges between them are represented by the dependency relation, e.g., nominal subject, adverbial modifier, etc. Given a neighborhood of a node $\mathcal{N}_i$, the node embeddings can be iteratively updated using multi-head attention as 
\begin{equation}
    h^{l+1}_{{att}_i} = concat_{k=1}^K \sum_{j \in \mathcal{N}_i} \alpha_{ij}^{lk} W_k^l h_j^l, 
\end{equation}
\begin{equation}
    \alpha_{ij}^{lk} = attention(i, j), 
\end{equation}
where $h^{l+1}_{{att}_i}$ is the attention head of node-$i$ at layer $l+1$ and $\alpha_{ij}^{lk}$ is the normalized attention coefficient computed by the $k$-th attention at layer $l$ and $W^l_k$ is an input transformation matrix.

In addition to the attention head of word-$i$ a relational head is also computed for this node as 
\begin{equation}
    h^{l+1}_{{rel}_i} = concat_{m=1}^M \sum_{j \in \mathcal{N}_i} \beta_{ij}^{lm} W_m^l h_j^l, 
\end{equation}
\begin{equation}
    g_{ij}^{lm} = \sigma (relu \left( r_{ij}W_{m1} + b_{m1}\right)W_{m2} + b_{m2})
\end{equation}
\begin{equation}
    \beta_{ij}^{lm} = exp(g_{ij}^{lm})/\sum_{j \in \mathcal{N}_i}exp(g_{ij}^{lm})
\end{equation}
where $r_{ij}$ denotes the relation embedding between node-$i$ and $j$. The final representation of each word (node) is a concatenation of the attention and relational embeddings:
\begin{equation}
    x_i^{l+1} = concat(h^{l+1}_{{att}_i}, h^{l+1}_{{rel}_i})
\end{equation}
\begin{equation}
    h^{l+1}_i = relu \left( W_{l+1}x_i^{l+1} + b_{l+1} \right)
\end{equation}

\subsection{Named Entity Recognition}
While the R-GAT model utilizes only the root representation to predict the sentiment polarity, here, we use the node representation $h^{l+1}_i$ for node-$i$ to predict the corresponding NER tag. 

\subsubsection{Linear Layer} In this case, $h^{l+1}_i$
is directly fed into a fully connected layer with softmax activation function that generates probabilities over all the NER classes, $\mathbf{C}$ 
\begin{equation}
    p(y_i = \mathbf{C}) = softmax(W_ph^{l+1}_i + b_p).
    \label{eq:prob}
\end{equation}
where $y_i$ is NER target label for the word-$i$ and $W_p, b_p$ are learnable parameters. 

\subsubsection{RNN Layer} While \citet{li-etal-2019-exploiting} applied GRU we have used a (Bi-directional Long Short Term Memory) BiLSTM layer on $h^{l+1}_i$ to capture the contributions of the neighbors of each token. An LSTM layer updates the input $h^{l+1}$ as 
\begin{align}
    i_t = \sigma\left( W_i[x_t;h_{t-1}] + b_i\right), \nonumber \\
    f_t = \sigma\left( W_f[x_t;h_{t-1}] + b_f\right), \nonumber \\
    o_t = \sigma\left( W_o[x_t;h_{t-1}] + b_o\right), \nonumber \\
    \tilde{c}_t = \tanh\left( W_c[x_t;h_{t-1}] + b_c\right), \nonumber \\
    c_t = f_t \odot c_{t-1} + i_t \odot \tilde{c}_t, \nonumber \\
    h_t = o_t \odot \tanh(c_t),  \nonumber
\end{align}
where $\odot$ denotes the Hadamard product, $W_i, W_f, W_o, W_c, b_i, b_f, b_o$ and $b_c$ are learnable parameters. The input to the LSTM layer $x_t = h^{l+1}_t$ and the output is $h^{l+2}_t$. A Bi-LSTM layer processes both the original sequence ($\overrightarrow{x}$) and the sequence in reverse order ($\overleftarrow{x}$) together so that contexts from both left and right are captured for each token. The final representation of each token is obtained by concatenating the outputs, i.e., $h^{l+2} = [\overrightarrow{h};\overleftarrow{h}]$ where $\overrightarrow{h} = LSTM(\overrightarrow{h}^{l+1})$ and $\overleftarrow{h} = LSTM(\overleftarrow{h}^{l+1})$.  

\subsubsection{Transformer Layer}
While an LSTM layer processes the sequence uni-directionally and the effects of neighboring tokens are accumulated one at a time, a Transformer \citep{vaswani2017attention} computes attention between all tokens irrespective of the direction. We apply a Transformer layer that takes $h^{l+1}$ as input and generates another sequence of token representations, $h^{l+2}$, which is of the same length.  

\subsubsection{CRF Layer}
For all the three previous cases, we have a final layer that generates a score matrix $\mathbf{P}$ over all the tag classes, where $P_{ij}$ is the score of the $j-$th tag for the $i-$th input token. The CRF layer learns a transition probability matrix $A \in \mathbb{R}^{K+2 \times K+2}$ where $K$ is the number of tag classes (and +2 indicates one tag each for the start and the end marker). For an input sequence $\mathbf{x}$ of length $T$ and the corresponding tags $\mathbf{y} = \{y_1, y_2, \ldots, y_T \}$ (with the start and end tag denoted by $y_0$ and $y_{T+1}$, the probability score is modified as
\begin{equation}
    S(\mathbf{x}, \mathbf{y}) = \sum_{i=0}^T A_{y_iy_{i+1}} + \sum_{i=1}^T P_{i, y_i}  
\end{equation}
which is used to calculate the probability of the tag sequence as 
\begin{equation}
    p(\mathbf{y} | \mathbf{x}) = e^{S(\mathbf{x}, \mathbf{y})}/\sum_{y' \in \mathbf{y}}e^{S(\mathbf{x}, \mathbf{y'})}
\end{equation}
which replaces Eq.~\ref{eq:prob}. The loss for each word is calculated by the categorical cross-entropy loss
\begin{equation}
    \mathcal{L}(\theta) = -\sum_{S \in \mathcal{D}}\sum_{w_i \in S}\log(p_{ij})
\end{equation}
where $p_{ij}$ is the probability of the word-$i$ with label-$j$ where the word-$i$ appeared in sentence-$S$ and $\mathcal{D}$ represents the entire corpus. For CRF based models, the loss is calculated using the forward-backward algorithm. 

\begin{table*}
\centering
\begin{tabular}{r|ccc|ccc}
\hline
Model & \multicolumn{3}{c}{Restaurant'14} & \multicolumn{3}{c}{Laptop'14} \\
 & P & R & F1 & P & R & F1 \\
\hline
\hline
RINANTE & 48.97 & 47.36 & 48.15 & 41.20 & 33.20 & 36.70 \\
CMLA & 67.80 & 73.69 & 70.62 & 54.70 & 59.20 & 56.90 \\
Li-unified & 74.43 & 69.26 & 71.75 & 68.01 & 56.72 & 61.86 \\
Li-unified-R & 73.15 & 74.44 & 73.79 & 66.28 & 60.71 & 63.38 \\
\citep{li-etal-2019-exploiting}–BLSTM & 70.00 & 74.20 & 72.04 & 65.99 & 54.62 & 59.77 \\
\citep{li-etal-2019-exploiting}–TG & 74.41 & 73.97 & 74.19 & 64.35 & 60.29 & 62.26 \\
\citep{li-etal-2019-exploiting}–T & 69.42 & 72.2 & 70.79 & 64.14 & 60.63 & 62.34 \\
\citep{li-etal-2019-exploiting} & 76.60 & 67.84 & 71.95 & 63.15 & 61.55 & 62.34 \\
\hline
RGAT-BERT & 71.70 & 78.80 & 75.08 & 60.69 & 62.74 & 61.69 \\
RGAT-BERT-CRF & 82.69 & 78.21 & \underline{80.39} & 70.38 & 62.53 & 66.22 \\
RGAT-BERT-BiLSTM-CRF & 81.20 & 79.39 & 80.28 & 70.43 & 68.21 & \textbf{69.30} \\
RGAT-BERT-TRFMR-CRF & 86.04 & 78.44 & \textbf{82.07} & 68.94 & 68.21 & \underline{68.57} \\
\hline
\end{tabular}
\caption{Comparison of predictions on the aspect extraction dataset of \cite{peng-et-al-2020-knowing}. The best F1-scores are shown in bold and the second best ones are underlined.}
\label{tab:astev1_results_p1}
\end{table*}

\begin{table*}
\centering
\begin{tabular}{r|ccc|ccc}
\hline
Model & \multicolumn{3}{c}{Restaurant'15} & \multicolumn{3}{c}{Restaurant'16} \\
 & P & R & F1 & P & R & F1 \\
\hline
\hline
RINANTE & 46.20 & 37.40 & 41.30 & 49.40 & 36.70 & 42.10 \\
CMLA & 49.90 & 58.00 & 53.60 & 58.90 & 63.60 & 61.20 \\
Li-unified & 61.39 & 67.99 & 64.52 & 66.88 & 71.40 & 69.06 \\
Li-unified-R & 64.95 & 64.95 & 64.95 & 66.33 & 74.55 & 70.20 \\
\citep{li-etal-2019-exploiting}–BLSTM & 63.41 & 65.19 & 64.29 & 69.74 & 71.62 & 70.67 \\
\citep{li-etal-2019-exploiting}–TG & 59.28 & 61.92 & 60.57 & 64.57 & 66.89 & 65.71 \\
\citep{li-etal-2019-exploiting}–T & 62.28 & 66.35 & 64.25 & 62.65 & 71.4 & 66.74 \\
\citep{li-etal-2019-exploiting} & 67.65 & 64.02 & 65.79 & 71.18 & 72.30 & 71.73 \\
\hline
RGAT-BERT & 62.37 & 71.59 & 66.67 & 66.35 & 78.15 & 71.77 \\
RGAT-BERT-CRF & 73.10 & 70.19 & 71.62 & 84.63 & 80.63 & \underline{82.58} \\
RGAT-BERT-BiLSTM-CRF & 76.03 & 73.71 & \textbf{74.85} & 84.61 & 81.76 & \textbf{83.16} \\
RGAT-BERT-TRFMR-CRF & 75.13 & 68.78 & \underline{71.81} & 84.36 & 80.18 & 82.22 \\
\hline
\end{tabular}
\caption{\label{astev1_results_p2}Comparison of predictions on the aspect extraction dataset of \cite{peng-et-al-2020-knowing}. The best F1-scores are shown in bold and the second best ones are underlined.}
\label{tab:astev1_results_p2}
\end{table*}

\section{Experiments}
\subsection{Datasets}
We used two datasets in our experiments. The first one is created by \citet{li-etal-2019-exploiting}, which is originating from SemEval'14 \citep{pontiki-etal-2014-semeval} but modified by \citet{li-etal-2019-e2ebert}. The statistics of the dataset are summarized in Table~\ref{astev0_data} where the number of sentences (queries) and aspects are shown in two domains, namely, Laptop and Restaurant, across train, validation and test set. 
The second dataset is called Aspect Sentiment Triplet Extraction (ASTE) dataset (version-1) as created by \citet{Peng_Xu_Bing_Huang_Lu_Si_2020} where each sentence has a unified aspect/target tags and opinion tags. The details of the dataset are shown in Table~\ref{tab:dataset2}.  

\begin{table*}
\centering
\begin{tabular}{r|ccc|ccc}
\hline
Model & \multicolumn{3}{c}{Restaurant'14} & \multicolumn{3}{c}{Laptop'14} \\
 & P & R & F1 & P & R & F1 \\
\hline
\hline
Distance rule & 58.39 & 43.59 & 49.92 & 50.13 & 33.86 & 40.42 \\
Dependency rule & 64.57 & 52.72 & 58.04 & 45.09 & 31.57 & 37.14 \\
RINANTE & 81.06 & 72.05 & 76.29 & 78.20 & 62.70 & 69.60 \\
CMLA & 69.47 & 74.53 & 71.91 & 51.80 & 65.30 & 57.70 \\
IOG & 82.85 & 77.38 & 80.02 & 73.24 & 69.63 & 71.35 \\
Li-unified-R & 81.20 & 83.18 & 82.13 & 76.62 & 74.90 & 75.70 \\
\citep{li-etal-2019-exploiting}–BLSTM & 80.41 & 86.19 & 83.15 & 78.06 & 68.98 & 73.19 \\
\citep{li-etal-2019-exploiting}–TG & 81.77 & 84.80 & 83.21 & 76.87 & 75.31 & 76.03 \\
\citep{li-etal-2019-exploiting}–T & 80.61 & 85.38 & 82.88 & 76.69 & 73.88 & 75.21 \\
\citep{li-etal-2019-exploiting} & 84.72 & 80.39 & 82.45 & 78.22 & 71.84 & 74.84 \\
\hline
RGAT-BERT & 82.05 & 86.43 & 84.18 & 74.71 & 80.20 & 77.36 \\
RGAT-BERT-CRF & 95.43 & 89.56 & \underline{92.40} & 93.97 & 79.59 & 86.19 \\
RGAT-BERT-BiLSTM-CRF & 95.64 & 91.53 & \textbf{93.54} & 93.42 & 84.08 & \textbf{88.51} \\
RGAT-BERT-TRFMR-CRF & 94.58 & 89.09 & 91.76 & 93.55 & 82.86 & \underline{87.88} \\
\hline
\end{tabular}
\caption{Comparison of predictions on the opinion extraction dataset of \citep{peng-et-al-2020-knowing} based on the SemEval'14 \citep{pontiki-etal-2014-semeval} task.}
\label{tab:astev1_results_o1}
\end{table*}

\subsection{Implementation Details}
We use the bi-affine parser \citep{dozat-manning-2016} from AllenNLP for dependency parsing. While \citet{wang-etal-2020-relational} used the aspect words to orient the dependency tree, in our case, we cannot use that information. Instead, we randomly choose a noun word (adjective for opinion tagging) from the input sentence, if available, otherwise, we select the middle token about which the dependency tree is re-oriented. For all experiments, the embedding dimension for the dependency relation is set to 200 and the dropout is fixed at 0.3. The last hidden state of the pre-trained BERT\footnote{https://github.com/huggingface/transformers} is used for the initial token representations which is subsequently fine-tuned. All models are trained using Adam optimizer \citep{kingma-adam-2014} with the default parameters. For all experiments we have used RGAT based feature extraction and BERT based token encoding. There are four variants of our model, namely, (1) RGAT-BERT, that does not use any other layer, (2) RGAT-BERT-CRF, that additionally uses a CRF final layer, (3) RGAT-BERT-BiLSTM-CRF, that uses a Bi-LSTM layer on the output of BERT before passing the output to a CRF layer and (4) RGAT-BERT-TRFMR-CRF that uses a Transformer layer instead of a Bi-LSTM. 

\subsection{Results and Discussions}
For all experiments we report only the F1-score (and omit precison and recall due to space limitation) across all the tags. Table~\ref{tab:perf-astev0} shows the performance of different RGAT models on the dataset of \citet{li-etal-2019-exploiting}. As can be seen, RGAT models provide the best F1 score for the Laptop (BiLSTM) and Restaurant (Transformer) domain surpassing the previous best F1 scores by more than 3\% point. For the laptop domain, the second best result is also given by another RGAT model (Transformer based), whereas, for the Restaurant domain one of the prior models stand out (BERT+self-attention network from \citet{li-etal-2019-exploiting}). 

For the second dataset, we extract aspects and opinions from four domains, Laptop, Restaurant-2014, Restaurant-2015 and Restaurant-2016. Table~\ref{tab:astev1_results_p1} and \ref{tab:astev1_results_p2} together show the results of aspect extraction for all the four domains. For aspect extraction, it can be seen that the best F1-score is obtained by the Transformer and BiLSTM models for the Restaurant-2014 and Laptop domain, respectively. The second best results are also obtained by our models, by CRF and Transformer based models. We can see substantial improvement (6-8\% point) over the previously reported F1 scores for both the cases. Even larger improvement can be seen for the other two Restaurant domain data (2015 and 2016) where the BiLSTM-CRF model generates the best F1 score.

Table~\ref{tab:astev1_results_o1} shows the performance on the opinion extraction task where the best F1 scores are provided by the BiLSTM-CRF model for both the Restaurant-2014 and Laptop domain. Here we see 10-12 \% point improvement over the best previous F1 score. Following the same trend of aspect extraction, for Restaurant-2015 and 2016 data (shown in Table~\ref{tab:astev1_results_o2}), we see the best performance coming from the BiLSTM-CRF model exceeding the previous best scores by more than 10\% point. Overall, we can see that all the RGAT-BERT based models perform much better than all the other baseline models on all the tasks.

\begin{table*}
\centering
\begin{tabular}{r|ccc|ccc}
\hline
Model & \multicolumn{3}{c}{Restaurant'15} & \multicolumn{3}{c}{Restaurant'16} \\
 & P & R & F1 & P & R & F1 \\
\hline
\hline
Distance rule & 54.12 & 39.96 & 45.97 & 61.90 & 44.57 & 51.83 \\
Dependency rule & 65.49 & 48.88 & 55.98 & 76.03 & 56.19 & 64.62 \\
RINANTE & 77.40 & 57.00 & 65.70 & 75.00 & 42.40 & 54.10 \\
CMLA & 60.80 & 65.30 & 62.90 & 74.50 & 69.00 & 71.70 \\
IOG & 76.06 & 70.71 & 73.25 & 85.25 & 78.51 & 81.69 \\
Li-unified-R & 79.18 & 75.88 & 77.44 & 79.84 & 86.88 & 83.16 \\
\citep{li-etal-2019-exploiting}–BLSTM & 74.29 & 80.48 & 77.21 & 82.12 & 84.95 & 83.46 \\
\citep{li-etal-2019-exploiting}–TG & 75.98 & 76.32 & 76.10 & 82.33 & 85.16 & 83.67 \\
\citep{li-etal-2019-exploiting}–T & 78.13 & 75.22 & 76.60 & 77.14 & 87.10 & 81.77 \\
\citep{li-etal-2019-exploiting} & 78.07 & 78.07 & 78.02 & 81.09 & 86.67 & 83.73 \\
\hline
RGAT-BERT & 75.20 & 81.98 & 78.44 & 77.98 & 89.89 & 83.52 \\
RGAT-BERT-CRF & 92.27 & 83.96 & \underline{87.92} & 96.07 & 84.09 & \underline{89.68} \\
RGAT-BERT-BiLSTM-CRF & 94.52 & 90.99 & \textbf{92.72} & 95.82 & 93.76 & \textbf{94.78} \\
RGAT-BERT-TRFMR-CRF & 93.62 & 80.66 & 86.66 & 94.37 & 83.01 & 88.33 \\
\hline
\end{tabular}
\caption{Comparison of predictions on the opinion extraction dataset of \citep{peng-et-al-2020-knowing} for the Restaurant domain based on \citealp{pontiki-etal-2015-semeval} and \citealp{pontiki-etal-2016-semeval}.}
\label{tab:astev1_results_o2}
\end{table*}

\section{Conclusion}
In this work we have applied Relational Graph Attention network which was previously used for classifying sentiment polarity associated with a specific aspect. However, here we show that the dependency structure of the input query when encoded by a Graph attention network is powerful enough to improve the performance and it is not necessary to tie to a particular aspect. We have used a surrogate aspect/opinion term by selecting a noun/adjective token if available. The strength of the approach is evident in the superior results that we have obtained for both aspect and opinion term extraction on four commonly used datasets. In addition, we have compared BiLSTM and Transformer as additional layers along with a CRF layer and found that BiLSTM layer consistently performs better than a Transformer layer. 

\section*{Limitations}
There are several limitations of the present methodology as shown below: 
\begin{enumerate}
    \item The success of the current approach heavily hinges on knowing the dependency structure of the queries. Thus, we cannot extend this approach easily to other languages other than English.
    
    \item While we have extracted all the aspects and opinions from a given query we have not matched them in this work. A more complete solution is proposed by \citet{mukherjee-etal-2021-paste} using pointer network based decoder, which can be the next step of this work. 
\end{enumerate}


\bibliographystyle{acl_natbib}
\bibliography{references}

\begin{thebibliography}{34}
\expandafter\ifx\csname natexlab\endcsname\relax\def\natexlab#1{#1}\fi

\bibitem[{Chen et~al.(2017)Chen, Sun, Bing, and Yang}]{chen-etal-2017-recurrent}
Peng Chen, Zhongqian Sun, Lidong Bing, and Wei Yang. 2017.
\newblock \href {https://doi.org/10.18653/v1/D17-1047} {Recurrent attention network on memory for aspect sentiment analysis}.
\newblock In \emph{Proceedings of the 2017 Conference on Empirical Methods in Natural Language Processing}, pages 452--461, Copenhagen, Denmark. Association for Computational Linguistics.

\bibitem[{Devlin et~al.(2019)Devlin, Chang, Lee, and Toutanova}]{devlin-etal-2019-bert}
Jacob Devlin, Ming-Wei Chang, Kenton Lee, and Kristina Toutanova. 2019.
\newblock \href {https://doi.org/10.18653/v1/N19-1423} {{BERT}: Pre-training of deep bidirectional transformers for language understanding}.
\newblock In \emph{Proceedings of the 2019 Conference of the North {A}merican Chapter of the Association for Computational Linguistics: Human Language Technologies, Volume 1 (Long and Short Papers)}, pages 4171--4186, Minneapolis, Minnesota. Association for Computational Linguistics.

\bibitem[{Dozat and Manning(2016)}]{dozat-manning-2016}
Timothy Dozat and Christopher~D Manning. 2016.
\newblock \href {http://arxiv.org/abs/1611.01734} {Deep biaffine attention for neural dependency parsing}.

\bibitem[{He et~al.(2018)He, Lee, Ng, and Dahlmeier}]{he-etal-2018-effective}
Ruidan He, Wee~Sun Lee, Hwee~Tou Ng, and Daniel Dahlmeier. 2018.
\newblock \href {https://aclanthology.org/C18-1096} {Effective attention modeling for aspect-level sentiment classification}.
\newblock In \emph{Proceedings of the 27th International Conference on Computational Linguistics}, pages 1121--1131, Santa Fe, New Mexico, USA. Association for Computational Linguistics.

\bibitem[{He et~al.(2019)He, Lee, Ng, and Dahlmeier}]{he-etal-2019-interactive}
Ruidan He, Wee~Sun Lee, Hwee~Tou Ng, and Daniel Dahlmeier. 2019.
\newblock \href {https://doi.org/10.18653/v1/P19-1048} {An interactive multi-task learning network for end-to-end aspect-based sentiment analysis}.
\newblock In \emph{Proceedings of the 57th Annual Meeting of the Association for Computational Linguistics}, pages 504--515, Florence, Italy. Association for Computational Linguistics.

\bibitem[{Hoang et~al.(2019)Hoang, Bihorac, and Rouces}]{hoang-etal-2019-aspect}
Mickel Hoang, Oskar~Alija Bihorac, and Jacobo Rouces. 2019.
\newblock \href {https://aclanthology.org/W19-6120} {Aspect-based sentiment analysis using {BERT}}.
\newblock In \emph{Proceedings of the 22nd Nordic Conference on Computational Linguistics}, pages 187--196, Turku, Finland. Link{\"o}ping University Electronic Press.

\bibitem[{Huang and Carley(2019)}]{huang-carley-2019-syntax}
Binxuan Huang and Kathleen Carley. 2019.
\newblock \href {https://doi.org/10.18653/v1/D19-1549} {Syntax-aware aspect level sentiment classification with graph attention networks}.
\newblock In \emph{Proceedings of the 2019 Conference on Empirical Methods in Natural Language Processing and the 9th International Joint Conference on Natural Language Processing (EMNLP-IJCNLP)}, pages 5469--5477, Hong Kong, China. Association for Computational Linguistics.

\bibitem[{Kingma and Ba(2014)}]{kingma-adam-2014}
Diederik~P. Kingma and Jimmy Ba. 2014.
\newblock \href {https://doi.org/10.48550/ARXIV.1412.6980} {Adam: A method for stochastic optimization}.

\bibitem[{Lample et~al.(2016)Lample, Ballesteros, Subramanian, Kawakami, and Dyer}]{lample-etal-2016-neural}
Guillaume Lample, Miguel Ballesteros, Sandeep Subramanian, Kazuya Kawakami, and Chris Dyer. 2016.
\newblock \href {https://doi.org/10.18653/v1/N16-1030} {Neural architectures for named entity recognition}.
\newblock In \emph{Proceedings of the 2016 Conference of the North {A}merican Chapter of the Association for Computational Linguistics: Human Language Technologies}, pages 260--270, San Diego, California. Association for Computational Linguistics.

\bibitem[{Li et~al.(2019{\natexlab{a}})Li, Bing, Li, and Lam}]{li-et-al-2019-unified}
Xin Li, Lidong Bing, Piji Li, and Wai Lam. 2019{\natexlab{a}}.
\newblock A unified model for opinion target extraction and target sentiment prediction.
\newblock In \emph{AAAI}, pages 6714–--6721.

\bibitem[{Li et~al.(2019{\natexlab{b}})Li, Bing, Li, and Lam}]{li-etal-2019-e2ebert}
Xin Li, Lidong Bing, Piji Li, and Wai Lam. 2019{\natexlab{b}}.
\newblock A unified model for opinion target extraction and target sentiment prediction.
\newblock In \emph{AAAI}, pages 6714--6721.

\bibitem[{Li et~al.(2019{\natexlab{c}})Li, Bing, Zhang, and Lam}]{li-etal-2019-exploiting}
Xin Li, Lidong Bing, Wenxuan Zhang, and Wai Lam. 2019{\natexlab{c}}.
\newblock \href {https://doi.org/10.18653/v1/D19-5505} {Exploiting {BERT} for end-to-end aspect-based sentiment analysis}.
\newblock In \emph{Proceedings of the 5th Workshop on Noisy User-generated Text (W-NUT 2019)}, pages 34--41, Hong Kong, China. Association for Computational Linguistics.

\bibitem[{Lin et~al.(2019)Lin, Yang, and Lai}]{ijcai2019-707}
Peiqin Lin, Meng Yang, and Jianhuang Lai. 2019.
\newblock \href {https://doi.org/10.24963/ijcai.2019/707} {Deep mask memory network with semantic dependency and context moment for aspect level sentiment classification}.
\newblock In \emph{Proceedings of the Twenty-Eighth International Joint Conference on Artificial Intelligence, {IJCAI-19}}, pages 5088--5094. International Joint Conferences on Artificial Intelligence Organization.

\bibitem[{Liu et~al.(2018)Liu, Shang, Ren, Xu, Gui, Peng, and Han.}]{liu-et-al-2018-empower}
Liyuan Liu, Jingbo Shang, Xiang Ren, Frank~F Xu, Huan Gui, Jian Peng, and Jiawei Han. 2018.
\newblock Empower sequence labeling with task-aware neural language model.
\newblock In \emph{AAAI}, pages 5253–--5260.

\bibitem[{Luo et~al.(2019)Luo, Li, Liu, and Zhang}]{luo-etal-2019-doer}
Huaishao Luo, Tianrui Li, Bing Liu, and Junbo Zhang. 2019.
\newblock \href {https://doi.org/10.18653/v1/P19-1056} {{DOER}: Dual cross-shared {RNN} for aspect term-polarity co-extraction}.
\newblock In \emph{Proceedings of the 57th Annual Meeting of the Association for Computational Linguistics}, pages 591--601, Florence, Italy. Association for Computational Linguistics.

\bibitem[{Ma et~al.(2017)Ma, Li, Zhang, and Wang}]{ijcai2017-568}
Dehong Ma, Sujian Li, Xiaodong Zhang, and Houfeng Wang. 2017.
\newblock \href {https://doi.org/10.24963/ijcai.2017/568} {Interactive attention networks for aspect-level sentiment classification}.
\newblock In \emph{Proceedings of the Twenty-Sixth International Joint Conference on Artificial Intelligence, {IJCAI-17}}, pages 4068--4074.

\bibitem[{Ma and Hovy(2016)}]{ma-hovy-2016-end}
Xuezhe Ma and Eduard Hovy. 2016.
\newblock \href {https://doi.org/10.18653/v1/P16-1101} {End-to-end sequence labeling via bi-directional {LSTM}-{CNN}s-{CRF}}.
\newblock In \emph{Proceedings of the 54th Annual Meeting of the Association for Computational Linguistics (Volume 1: Long Papers)}, pages 1064--1074, Berlin, Germany. Association for Computational Linguistics.

\bibitem[{Mukherjee et~al.(2021)Mukherjee, Nayak, Butala, Bhattacharya, and Goyal}]{mukherjee-etal-2021-paste}
Rajdeep Mukherjee, Tapas Nayak, Yash Butala, Sourangshu Bhattacharya, and Pawan Goyal. 2021.
\newblock \href {https://doi.org/10.18653/v1/2021.emnlp-main.731} {{PASTE}: A tagging-free decoding framework using pointer networks for aspect sentiment triplet extraction}.
\newblock In \emph{Proceedings of the 2021 Conference on Empirical Methods in Natural Language Processing}, pages 9279--9291, Online and Punta Cana, Dominican Republic. Association for Computational Linguistics.

\bibitem[{Peng et~al.(2020{\natexlab{a}})Peng, Xu, Bing, Huang, Lu, and Si}]{Peng_Xu_Bing_Huang_Lu_Si_2020}
Haiyun Peng, Lu~Xu, Lidong Bing, Fei Huang, Wei Lu, and Luo Si. 2020{\natexlab{a}}.
\newblock \href {https://doi.org/10.1609/aaai.v34i05.6383} {Knowing what, how and why: A near complete solution for aspect-based sentiment analysis}.
\newblock \emph{Proceedings of the AAAI Conference on Artificial Intelligence}, 34(05):8600--8607.

\bibitem[{Peng et~al.(2020{\natexlab{b}})Peng, Xu, Bing, Huang, Lu, and Si}]{peng-et-al-2020-knowing}
Haiyun Peng, Lu~Xu, Lidong Bing, Fei Huang, Wei Lu, and Luo Si. 2020{\natexlab{b}}.
\newblock Knowing what, how and why: A near complete solution for aspect-based sentiment analysis.
\newblock In \emph{AAAI}, volume 34 (5).

\bibitem[{Pontiki et~al.(2016)Pontiki, Galanis, Papageorgiou, Androutsopoulos, Manandhar, AL-Smadi, Al-Ayyoub, Zhao, Qin, De~Clercq, Hoste, Apidianaki, Tannier, Loukachevitch, Kotelnikov, Bel, Jim{\'e}nez-Zafra, and Eryi{\u{g}}it}]{pontiki-etal-2016-semeval}
Maria Pontiki, Dimitris Galanis, Haris Papageorgiou, Ion Androutsopoulos, Suresh Manandhar, Mohammad AL-Smadi, Mahmoud Al-Ayyoub, Yanyan Zhao, Bing Qin, Orph{\'e}e De~Clercq, V{\'e}ronique Hoste, Marianna Apidianaki, Xavier Tannier, Natalia Loukachevitch, Evgeniy Kotelnikov, Nuria Bel, Salud~Mar{\'\i}a Jim{\'e}nez-Zafra, and G{\"u}l{\c{s}}en Eryi{\u{g}}it. 2016.
\newblock \href {https://doi.org/10.18653/v1/S16-1002} {{S}em{E}val-2016 task 5: Aspect based sentiment analysis}.
\newblock In \emph{Proceedings of the 10th International Workshop on Semantic Evaluation ({S}em{E}val-2016)}, pages 19--30, San Diego, California. Association for Computational Linguistics.

\bibitem[{Pontiki et~al.(2015)Pontiki, Galanis, Papageorgiou, Manandhar, and Androutsopoulos}]{pontiki-etal-2015-semeval}
Maria Pontiki, Dimitris Galanis, Haris Papageorgiou, Suresh Manandhar, and Ion Androutsopoulos. 2015.
\newblock \href {https://doi.org/10.18653/v1/S15-2082} {{S}em{E}val-2015 task 12: Aspect based sentiment analysis}.
\newblock In \emph{Proceedings of the 9th International Workshop on Semantic Evaluation ({S}em{E}val 2015)}, pages 486--495, Denver, Colorado. Association for Computational Linguistics.

\bibitem[{Pontiki et~al.(2014)Pontiki, Galanis, Pavlopoulos, Papageorgiou, Androutsopoulos, and Manandhar}]{pontiki-etal-2014-semeval}
Maria Pontiki, Dimitris Galanis, John Pavlopoulos, Harris Papageorgiou, Ion Androutsopoulos, and Suresh Manandhar. 2014.
\newblock \href {https://doi.org/10.3115/v1/S14-2004} {{S}em{E}val-2014 task 4: Aspect based sentiment analysis}.
\newblock In \emph{Proceedings of the 8th International Workshop on Semantic Evaluation ({S}em{E}val 2014)}, pages 27--35, Dublin, Ireland. Association for Computational Linguistics.

\bibitem[{Sun et~al.(2019)Sun, Zhang, Mensah, Mao, and Liu}]{sun-etal-2019-aspect}
Kai Sun, Richong Zhang, Samuel Mensah, Yongyi Mao, and Xudong Liu. 2019.
\newblock \href {https://doi.org/10.18653/v1/D19-1569} {Aspect-level sentiment analysis via convolution over dependency tree}.
\newblock In \emph{Proceedings of the 2019 Conference on Empirical Methods in Natural Language Processing and the 9th International Joint Conference on Natural Language Processing (EMNLP-IJCNLP)}, pages 5679--5688, Hong Kong, China. Association for Computational Linguistics.

\bibitem[{Tang et~al.(2016)Tang, Qin, and Liu}]{tang-etal-2016-aspect}
Duyu Tang, Bing Qin, and Ting Liu. 2016.
\newblock \href {https://doi.org/10.18653/v1/D16-1021} {Aspect level sentiment classification with deep memory network}.
\newblock In \emph{Proceedings of the 2016 Conference on Empirical Methods in Natural Language Processing}, pages 214--224, Austin, Texas. Association for Computational Linguistics.

\bibitem[{Tay et~al.(2017)Tay, Tuan, and Hui}]{10.1145/3132847.3132936}
Yi~Tay, Luu~Anh Tuan, and Siu~Cheung Hui. 2017.
\newblock \href {https://doi.org/10.1145/3132847.3132936} {Dyadic memory networks for aspect-based sentiment analysis}.
\newblock In \emph{Proceedings of the 2017 ACM on Conference on Information and Knowledge Management}, CIKM '17, page 107–116, New York, NY, USA. Association for Computing Machinery.

\bibitem[{Vaswani et~al.(2017)Vaswani, Shazeer, Parmar, Uszkoreit, Jones, Gomez, Kaiser, and Polosukhin}]{vaswani2017attention}
Ashish Vaswani, Noam Shazeer, Niki Parmar, Jakob Uszkoreit, Llion Jones, Aidan~N. Gomez, Lukasz Kaiser, and Illia Polosukhin. 2017.
\newblock \href {http://arxiv.org/abs/1706.03762} {Attention is all you need}.

\bibitem[{Wang et~al.(2018)Wang, Li, Li, Kang, Zhang, Si, and Zhou}]{ijcai2018-617}
Jingjing Wang, Jie Li, Shoushan Li, Yangyang Kang, Min Zhang, Luo Si, and Guodong Zhou. 2018.
\newblock \href {https://doi.org/10.24963/ijcai.2018/617} {Aspect sentiment classification with both word-level and clause-level attention networks}.
\newblock In \emph{Proceedings of the Twenty-Seventh International Joint Conference on Artificial Intelligence, {IJCAI-18}}, pages 4439--4445. International Joint Conferences on Artificial Intelligence Organization.

\bibitem[{Wang et~al.(2020)Wang, Shen, Yang, Quan, and Wang}]{wang-etal-2020-relational}
Kai Wang, Weizhou Shen, Yunyi Yang, Xiaojun Quan, and Rui Wang. 2020.
\newblock \href {https://doi.org/10.18653/v1/2020.acl-main.295} {Relational graph attention network for aspect-based sentiment analysis}.
\newblock In \emph{Proceedings of the 58th Annual Meeting of the Association for Computational Linguistics}, pages 3229--3238, Online. Association for Computational Linguistics.

\bibitem[{Wang et~al.(2016)Wang, Huang, Zhu, and Zhao}]{wang-etal-2016-attention}
Yequan Wang, Minlie Huang, Xiaoyan Zhu, and Li~Zhao. 2016.
\newblock \href {https://doi.org/10.18653/v1/D16-1058} {Attention-based {LSTM} for aspect-level sentiment classification}.
\newblock In \emph{Proceedings of the 2016 Conference on Empirical Methods in Natural Language Processing}, pages 606--615, Austin, Texas. Association for Computational Linguistics.

\bibitem[{Xu et~al.(2019)Xu, Liu, Shu, and Yu}]{xu2019bert}
Hu~Xu, Bing Liu, Lei Shu, and Philip~S. Yu. 2019.
\newblock \href {http://arxiv.org/abs/1904.02232} {Bert post-training for review reading comprehension and aspect-based sentiment analysis}.

\bibitem[{Zeng et~al.(2019)Zeng, Yang, Xu, Zhou, and Han}]{app9163389}
Biqing Zeng, Heng Yang, Ruyang Xu, Wu~Zhou, and Xuli Han. 2019.
\newblock \href {https://doi.org/10.3390/app9163389} {Lcf: A local context focus mechanism for aspect-based sentiment classification}.
\newblock \emph{Applied Sciences}, 9(16).

\bibitem[{Zhang et~al.(2019)Zhang, Li, and Song}]{zhang-etal-2019-aspect}
Chen Zhang, Qiuchi Li, and Dawei Song. 2019.
\newblock \href {https://doi.org/10.18653/v1/D19-1464} {Aspect-based sentiment classification with aspect-specific graph convolutional networks}.
\newblock In \emph{Proceedings of the 2019 Conference on Empirical Methods in Natural Language Processing and the 9th International Joint Conference on Natural Language Processing (EMNLP-IJCNLP)}, pages 4568--4578, Hong Kong, China. Association for Computational Linguistics.

\bibitem[{Zhang et~al.(2020)Zhang, Li, Song, and Wang}]{zhang-etal-2020-multi-task}
Chen Zhang, Qiuchi Li, Dawei Song, and Benyou Wang. 2020.
\newblock \href {https://doi.org/10.18653/v1/2020.findings-emnlp.72} {A multi-task learning framework for opinion triplet extraction}.
\newblock In \emph{Findings of the Association for Computational Linguistics: EMNLP 2020}, pages 819--828, Online. Association for Computational Linguistics.

\end{thebibliography}

\end{document}